\documentclass[conference]{IEEEtran}
\IEEEoverridecommandlockouts

\usepackage{cite}
\usepackage{amsmath,amssymb,amsfonts}
\usepackage{algorithmic}
\usepackage{graphicx}
\usepackage{textcomp}
\usepackage{xcolor}
\usepackage{url}
\usepackage{booktabs}
\usepackage{multirow}
\usepackage{enumitem}

\graphicspath{{figures/}{../figures/}}

\def\BibTeX{{\rm B\kern-.05em{\sc i\kern-.025em b}\kern-.08em
    T\kern-.1667em\lower.7ex\hbox{E}\kern-.125emX}}

\newcommand{\tumAllLinesOff}{0.1960}
\newcommand{\tumAllLinesOn}{0.1205}
\newcommand{\tumAllDelta}{$-$38.5\%}
\newcommand{\tumAllMedianLinesOff}{0.1436}
\newcommand{\tumAllMedianLinesOn}{0.0511}
\newcommand{\tumAllMedianDelta}{$-$64.4\%}

\newcommand{\tumSubsetLinesOff}{0.1112}
\newcommand{\tumSubsetLinesOn}{0.0357}
\newcommand{\tumSubsetDelta}{$-$67.9\%}

\newcommand{\replicaMeanTransDelta}{$-$30.1\%}
\newcommand{\replicaMedianTransDelta}{$-$38.9\%}
\newcommand{\replicaMeanRotDelta}{$-$26.8\%}
\newcommand{\replicaMedianRotDelta}{$-$66.4\%}

\newcommand{\tumAllDeltaMag}{38.5\%}

\newcommand{\replicaMeanTransDeltaMag}{30.1\%}
\newcommand{\replicaMedianTransDeltaMag}{38.9\%}

\newcommand{\replicaMedianRotDeltaMag}{66.4\%}

\newcommand{\nostructFarDelta}{$-$74.1\%}

\newcommand{\largeCabinetDelta}{$-$79.2\%}

\providecommand{\tumAllLinesOff}{??}
\providecommand{\tumAllLinesOn}{??}
\providecommand{\tumAllDelta}{??}
\providecommand{\tumSubsetLinesOff}{??}
\providecommand{\tumSubsetLinesOn}{??}
\providecommand{\tumSubsetDelta}{??}
\providecommand{\replicaMeanTransDelta}{??}
\providecommand{\replicaMedianTransDelta}{??}
\providecommand{\replicaMeanRotDelta}{??}
\providecommand{\replicaMedianRotDelta}{??}

\providecommand{\nostructFarDelta}{??}

\providecommand{\largeCabinetDelta}{??}
\providecommand{\tumAllMedianLinesOff}{??}
\providecommand{\tumAllMedianLinesOn}{??}
\providecommand{\tumAllMedianDelta}{??}
\providecommand{\tumAllDeltaMag}{??}

\providecommand{\replicaMeanTransDeltaMag}{??}
\providecommand{\replicaMedianTransDeltaMag}{??}

\providecommand{\replicaMedianRotDeltaMag}{??}

\newcommand{\vn}{\mathbf{n}}
\newcommand{\vv}{\mathbf{v}}
\newcommand{\vL}{\mathcal{L}}
\newcommand{\skewm}[1]{\lfloor #1 \rfloor_\times}

\newcommand{\eqn}[1]{equation~\ref{#1}}
\newcommand{\eqns}[2]{equations~\ref{#1}--\ref{#2}}

\begin{document}

\title{Line-Augmented cuVSLAM: Robust Visual Odometry in Low-Texture Environments}

\author{ \IEEEauthorblockN{ Dara Molloy\IEEEauthorrefmark{1}, Terry Wu\IEEEauthorrefmark{1}, Wonju Lee\IEEEauthorrefmark{1}, Abhilash Nand Kumar\IEEEauthorrefmark{1}, David Adley\IEEEauthorrefmark{1}, \\ Conor Ferguson\IEEEauthorrefmark{1}, Hesam Rabeti\IEEEauthorrefmark{2} } \IEEEauthorblockA{ \IEEEauthorrefmark{1}\textit{Analog Devices} \qquad \IEEEauthorrefmark{2}\textit{NVIDIA}\\ Dara.Molloy@analog.com } }

\maketitle

\begin{abstract}
Autonomous mobile robots (AMRs) in warehouses and production facilities operate in
precisely the environments that defeat point-feature visual odometry: featureless
aisles, uniform walls, repetitive racking and specular floors. Such environments are,
however, rich in straight-edge structure: rack uprights, door frames, wall
junctions and cabinet outlines present high-contrast linear edges stable across wide
viewpoint changes. A line-feature extension to cuVSLAM~\cite{korovko_cuvslam_2025},
a CUDA-accelerated visual-SLAM system, is presented that recovers this
signal. Lines are parameterised as Pl\"{u}cker coordinates for geometric transport
and as the minimal 4-DOF orthonormal representation of Bartoli and Sturm~\cite{bartoli_sfm_lines_2005}
for optimisation; a Cauchy-robust point-to-line residual with analytic Jacobians is
folded directly into cuVSLAM's per-frame Gauss--Newton pose solve, and the same line
state is jointly refined in the Schur-complement sliding-window bundle adjustment. An
adaptive information weight causes lines to assume more of the estimation burden exactly
as point features become scarce, and the extension adds residual blocks to existing
solvers rather than introducing a second estimator, reducing exactly to the unmodified
point pipeline when disabled. Evaluated in RGB-D mode under SE(3) Umeyama alignment with
no scale correction, line features reduce mean absolute trajectory error by
\tumAllDeltaMag\ over the 22 sequences of the TUM RGB-D freiburg3 collection and by
\replicaMeanTransDeltaMag\ over the 8 Replica room and office sequences, where median
translation error falls by \replicaMedianTransDeltaMag\ and median rotation error by
\replicaMedianRotDeltaMag, with zero tracking failures in any configuration tested.
\end{abstract}

\section{Introduction}
\label{sec:intro}

Cameras are inexpensive, compact and information-dense, making visual odometry an
attractive alternative to LIDAR-based localisation for autonomous mobile robots,
provided it can match LIDAR's robustness in the industrial
environments where point-feature tracking degrades.
cuVSLAM~\cite{korovko_cuvslam_2025} represents a modern, state-of-the-art
visual-SLAM system: CUDA-accelerated, supporting up to 32 cameras across RGB,
depth and stereo-inertial pipelines, and running in real time on NVIDIA Jetson
edge hardware.
As in ORB-SLAM3~\cite{campos_orbslam3_2021}, VINS-Mono~\cite{qin_vinsmono_2018} and other
deployed visual or visual-inertial odometry (VIO) systems, its front end tracks sparse
point features: corners and textured patches carried across frames by descriptor
matching or optical-flow tracking. cuVSLAM's own point path uses the latter: GFTT
corner detection followed by GPU Lucas--Kanade optical-flow tracking with
normalised-cross-correlation validation, not descriptor re-matching.

Point features are precisely the primitive that indoor structured environments lack. A
warehouse aisle or production facility presents long stretches of uniformly painted wall,
repetitive shelving, specular flooring and low-contrast or distant ceiling structure. Corner
detectors yield few correspondences in these conditions, and the surviving matches are
often poorly conditioned or attached to moving inventory.

These environments are, however, rich in straight-edge structure. Rack uprights, door
frames, wall--floor junctions, pallet edges and cabinet outlines are abundant,
high-contrast and stable across wide viewpoint changes. A line is also a
better-conditioned measurement than a point in these scenes, as it aggregates gradient
evidence along its full length, which increases robustness to noise, motion blur and
partial occlusion relative to a small tracked point patch. The point-line VIO
literature converges on this observation, reporting that line features yield the
largest gains in degenerate, low-texture and structured
scenes~\cite{fu_plvins_2020,he_plvio_2018,zou_structvio_2018}.

This paper carries the point-line VIO formulation established by
PL-VINS~\cite{fu_plvins_2020} and its lineage into cuVSLAM's CUDA-accelerated pose
solve and bundle adjustment, and extends it with five departures aimed at the
industrial degradation regime above. The extension augments rather than replaces the
point pipeline and preserves the low-latency front end/back end separation that makes
cuVSLAM deployable on edge hardware. The contributions are:

\begin{itemize}
\item A \textbf{persistent, cross-window line landmark registry} 
that keeps a landmark alive after its originating frame
leaves the bundle-adjustment sliding window, so an early, wide-baseline observation
can still be paired against a much later one for triangulation, in place of
PL-VINS's window-scoped landmarks, which are deleted outright once their originating
frame marginalises out.

\item A \textbf{trimmed-mean multi-view line fusion} that
ranks each landmark's recently triangulated endpoint pairs by mutual agreement and
discards a fixed worst-agreeing fraction before averaging, in place of PL-VINS's
landmark state, which is otherwise a free variable updated only by the joint bundle
adjustment, with no separate fusion step.

\item An \textbf{adaptive, clamped point/line weight} that scales inversely with
surviving point observations and saturates at fixed floor/ceiling bounds, in 
place of PL-VINS's single fixed information matrix
shared unweighted between points and lines.

\item A \textbf{cuVSLAM integration of point-to-line residuals} into both the
per-frame front end pose solve, with a Cauchy robust loss, and the windowed back end
bundle adjustment, with a Student's-$t$ robust loss.

\item An \textbf{evaluation on two public benchmarks}, TUM RGB-D and Replica, 
under strict SE(3) alignment with no scale
correction, paired with an analysis of
which scene properties predict benefit.
\end{itemize}

\section{Related Work}
\label{sec:related-work}

Early visual-SLAM systems were formulated as recursive filters, with camera poses and map
features maintained in a state updated around a local linearisation
\cite{huang_observability_2010}. Nonlinear factor-graph optimization subsequently became
standard, with bundle adjustment providing a sparse pose-landmark formulation for joint
refinement~\cite{triggs_bundle_2000}. Keyframe-based systems such as ORB-SLAM3
\cite{campos_orbslam3_2021} established this point-centric paradigm across monocular,
stereo, RGB-D, and visual-inertial configurations. cuVSLAM~\cite{korovko_cuvslam_2025}
extends this lineage to CUDA-accelerated, multi-camera, real-time edge deployment.

Early point-line systems commonly represented a mapped segment by two 3D endpoints.
In the original PL-SLAM implementation~\cite{gomez_plslam_2017}, the endpoint
coordinates were optimized directly, and the residual was formed from the projected
endpoints and the corresponding observed infinite image line. This representation
simplified initialization and association, but endpoint locations remained sensitive to
occlusion, contrast variation, and partial visibility. Metric depth does not remove this
issue in RGB-D systems, because endpoint locations remain affected by image truncation.
Later formulations replaced endpoint coordinates with geometric line states while
retaining endpoint-to-line reprojection constraints, as in PL-VIO~\cite{he_plvio_2018}.

The geometric foundation was provided by the Pl\"{u}cker representation, which supports
rigid transport and projection but uses six coordinates subject to the Klein constraint.
A minimal four-degree-of-freedom orthonormal representation was established for line
reconstruction and bundle adjustment by Bartoli and Sturm
\cite{bartoli_sfm_lines_2005}. Pl\"{u}cker coordinates can consequently be retained for
transport while the orthonormal chart is used for optimization, avoiding the rank
deficiency of direct constrained optimization. This representation was incorporated into
tightly coupled point-line estimation by PL-VIO~\cite{he_plvio_2018} and PL-VINS
\cite{fu_plvins_2020}. Line measurements complement point measurements but do not remove
the underlying unobservable directions~\cite{observability_hetero_2018}.

The Pl\"{u}cker--orthonormal formulation has been implemented in visual-inertial estimation
\cite{zhang_plcvio_2023}, wheel-odometry-aided odometry
\cite{zhang_plviwo_2025}, and RGB-D SLAM \cite{sun_uplslam_2025}. Structural-prior methods 
such as StructVIO~\cite{zou_structvio_2018}
obtain additional constraints from Manhattan-world regularity; no such axis-alignment
assumption is imposed here.

In this work, the established line geometry is integrated into cuVSLAM's
CUDA-accelerated frontend and backend, with PL-VINS serving as the principal
architectural reference. Three limitations of the inherited design motivate corresponding
extensions for persistent line estimation in low-texture, structure-rich environments:

\begin{itemize}
\item \emph{Landmark persistence} Line landmarks are scoped to the active sliding
window. When the frame associated with a landmark is marginalized, its identity is not
retained for later re-association; a subsequent observation of the same edge is therefore
treated as a new landmark. Long-lived structural edges cannot consequently accumulate
observations across successive windows.

\item \emph{Multi-view reconstruction} In the inherited monocular formulation, a line
landmark is initialized from a selected observation pair and subsequently refined as a
free bundle-adjustment variable. No separate multi-view fusion step is applied, so an
erroneous association or poorly conditioned reconstruction can influence the landmark
state.

\item \emph{Residual weighting} Point and line residuals share a fixed information
matrix under a Huber norm~\cite{huber_robust_1992}, without a relative point-line
weight. Their relative authority therefore remains constant as point support varies.
\end{itemize}

\noindent Sections~\ref{sec:fusion} and~\ref{sec:weighting} address these three points in
turn.

\paragraph{Detection and description}
LSD~\cite{gioi_lsd_2012} is used for its subpixel localization and false-detection
control; EDLines~\cite{akinlar_edlines_2011} and ELSED~\cite{suarez_elsed_2022} provide
alternative accuracy--runtime trade-offs. Detection is exposed through an interchangeable
interface (Section~\ref{sec:detection}), and association is performed with the Line Band
Descriptor~\cite{zhang_lbd_2013}, as in the PL-VIO/PL-VINS lineage
\cite{he_plvio_2018,fu_plvins_2020}.

\section{System Overview}
\label{sec:system}

\begin{figure*}[t!]
\centering
\includegraphics[width=0.9\textwidth]{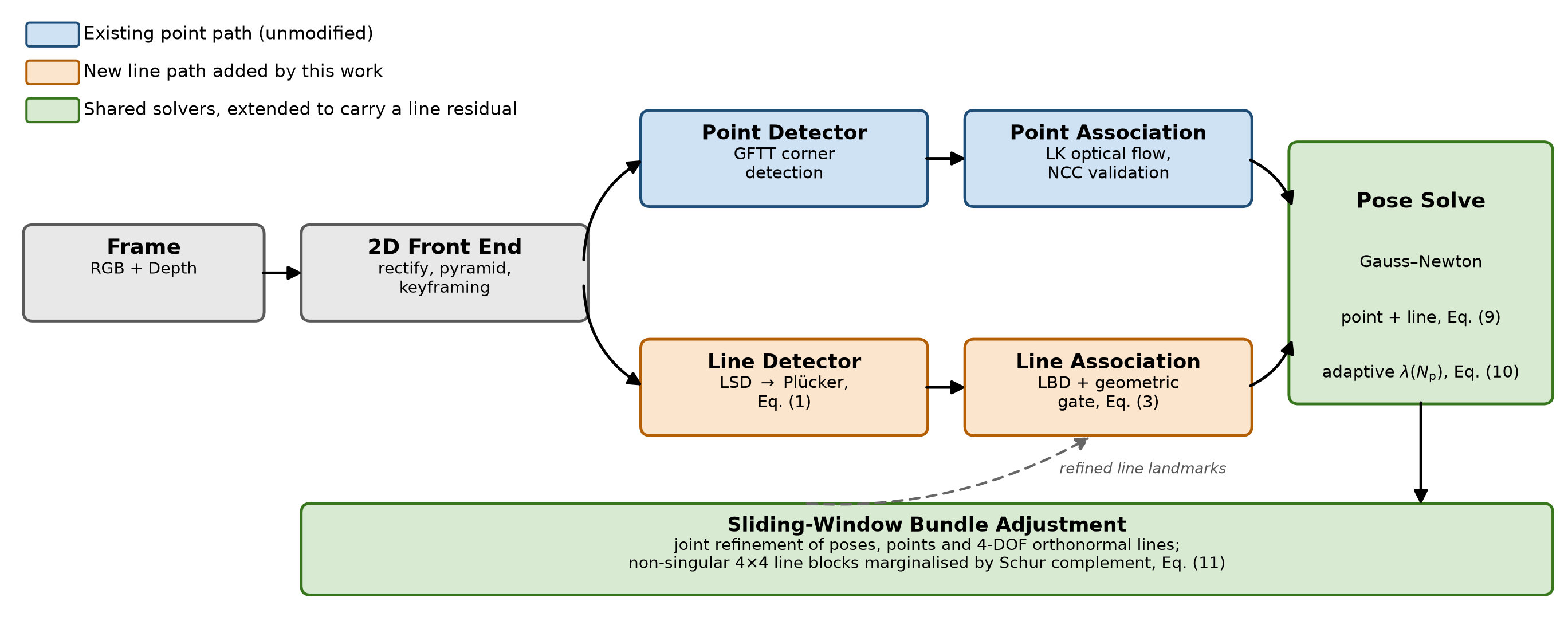}
\caption{Line-augmented cuVSLAM pipeline. The point path is unmodified; the line path
runs in parallel and converges on the same Gauss--Newton pose solve, where
$\lambda(N_{\mathrm{p}})$ of \eqn{eq:adaptive} shifts authority towards lines as
point observations become scarce, and on the same bundle adjustment, where the
minimal orthonormal parameterisation lets line landmarks marginalise by the same Schur
complement already used for points.}
\label{fig:architecture}
\end{figure*}

cuVSLAM~\cite{korovko_cuvslam_2025} separates a low-latency front end, which
tracks features and estimates the per-frame pose against a local odometry map, from an
asynchronous back end, which refines that map through sliding-window bundle
adjustment and maintains global consistency through pose-graph optimisation and loop
closure. This split enables real-time operation on Jetson-class hardware, and our
design preserves it without modification.

The line path attaches at both levels without altering the separation
(Fig.~\ref{fig:architecture}). A persistent line-landmark registry is maintained
outside the active bundle-adjustment state and stores each landmark's identity,
geometric state, descriptor, and recent depth-based reconstructions. Landmarks leaving
the sliding window can therefore be re-associated through geometric prediction and
descriptor matching, enabling their accumulated endpoint observations to reinforce current estimates. Their residuals
are incorporated into the existing Gauss--Newton pose solve and sliding-window bundle
adjustment. Schur marginalization thus removes only the active variable; the registry
entry is retained, and no parallel estimator or pose-graph arbitration is required.

\section{Line-Feature Formulation}
\label{sec:formulation}

We represent each 3D line with two parameterisations, selected for their respective
strengths: Pl\"{u}cker coordinates for rigid transport and projection, where the algebra
is linear, and the orthonormal representation for the optimisation state, where
minimality guarantees a well-conditioned normal equation. Conversions occur only at the
boundaries between these roles.

\subsection{Pl\"{u}cker Coordinates}
\label{sec:plucker}

A line through a point $\mathbf{p} \in \mathbb{R}^3$ with unit direction
$\vv \in \mathbb{R}^3$ is written~\cite{hartley_zisserman_2004}
\begin{equation}
\vL = \begin{bmatrix} \vn \\ \vv \end{bmatrix} \in \mathbb{R}^6,
\qquad \vn = \mathbf{p} \times \vv ,
\label{eq:plucker}
\end{equation}
where the moment $\vn$ is normal to the plane spanned by the line and the origin, and
$\|\vn\|/\|\vv\|$ is the origin-to-line distance. The coordinates are homogeneous and
satisfy the Klein quadric constraint $\vn^\top \vv = 0$, leaving four degrees of
freedom in six parameters.

The decisive property for line transport is that rigid motion acts linearly. For a
transform $\mathbf{T}_{cw} = (\mathbf{R}_{cw}, \mathbf{t}_{cw}) \in \mathrm{SE}(3)$
from world to camera,
\begin{equation}
\vL_c
=\underbrace{\begin{bmatrix}
\mathbf{R}_{cw} & \skewm{\mathbf{t}_{cw}}\mathbf{R}_{cw} \\
\mathbf{0}      & \mathbf{R}_{cw}
\end{bmatrix}}_{\mathcal{H}_{cw}} \vL_w ,
\label{eq:plk-transform}
\end{equation}
with $\skewm{\cdot}$ the skew-symmetric cross-product matrix, i.e.\ the linear
operator such that $\skewm{\mathbf{a}}\mathbf{b} = \mathbf{a}\times\mathbf{b}$ for
any $\mathbf{a}, \mathbf{b} \in \mathbb{R}^3$. Under a calibrated
pinhole camera the projected image line depends only on the transported moment,
\begin{equation}
\boldsymbol{\ell} = \begin{bmatrix} \ell_1 \\ \ell_2 \\ \ell_3 \end{bmatrix}
= \mathbf{K}_{\!\vL}\, \vn_c ,
\quad
\mathbf{K}_{\!\vL} =
\begin{bmatrix}
f_v & 0 & 0\\
0 & f_u & 0\\
-f_v c_u & -f_u c_v & f_u f_v
\end{bmatrix},
\label{eq:line-projection}
\end{equation}
where $\mathbf{K}_{\!\vL}$ is the line-projection matrix induced by the pinhole
intrinsics $(f_u, f_v, c_u, c_v)$. The direction $\vv_c$ drops out of the projection in
\eqn{eq:line-projection}, so a line observation constrains only two of the four line
degrees of freedom per view, and lines must be accumulated across views to be fully
determined. Each additional view contributes another rank-two endpoint-to-projected-line
constraint. Bundle adjustment jointly updates all four orthonormal line coordinates
through the residual Jacobian with respect to the tangent perturbation of
\eqn{eq:orth-update}, so views with sufficient angular diversity recover the remaining
degrees of freedom.

\subsection{Orthonormal Representation}
\label{sec:orthonormal}

Optimising the Pl\"{u}cker coordinates of \eqn{eq:plucker} directly is ill-posed: six
parameters carry four degrees of freedom, so the normal equations are rank-deficient by
two. The minimal
orthonormal representation of Bartoli and Sturm~\cite{bartoli_sfm_lines_2005} is adopted: every
Pl\"{u}cker line admits a unique
$(\mathbf{U}, \mathbf{W}) \in \mathrm{SO}(3) \times \mathrm{SO}(2)$ with
\begin{equation}
\mathbf{U} =
\begin{bmatrix}
\dfrac{\vn}{\|\vn\|} &
\dfrac{\vv}{\|\vv\|} &
\dfrac{\vn \times \vv}{\|\vn \times \vv\|}
\end{bmatrix},
\quad
\mathbf{W} =
\begin{bmatrix} w_1 & -w_2 \\ w_2 & w_1 \end{bmatrix},
\label{eq:orth}
\end{equation}
where $(w_1, w_2) = (\|\vn\|, \|\vv\|)/\sigma$, $\sigma = \sqrt{\|\vn\|^2 + \|\vv\|^2}$,
so $w_1^2 + w_2^2 = 1$, and the ratio $w_2/w_1$ encodes the origin-to-line distance,
degrading gracefully towards lines at infinity rather than becoming singular. Updates
are applied on the four-dimensional tangent space,
\begin{equation}
\mathbf{U} \leftarrow \mathbf{U}\,\mathrm{Exp}(\boldsymbol{\theta}), \qquad
\mathbf{W} \leftarrow \mathbf{W}\,\mathrm{Exp}(\phi),
\label{eq:orth-update}
\end{equation}
for $\boldsymbol{\delta} = [\,\boldsymbol{\theta}^\top,\, \phi\,]^\top \in \mathbb{R}^4$,
so the update is multiplicative, stays on the manifold, and the Klein constraint holds
exactly by construction~\cite{sola_micro_2018}.

\subsection{Observation Model and Jacobians}
\label{sec:residual}

A tracked 2D segment is observed through its two endpoints
$\mathbf{x}_s, \mathbf{x}_e \in \mathbb{P}^2$, expressed in normalised camera
coordinates (undistorted, focal-length- and principal-point-free). Rather than
penalise endpoint-to-endpoint distance, which is corrupted by the unstable segment
extent produced by occlusion, the perpendicular distance from each
endpoint to the infinite projected line is penalized, following PL-VIO~\cite{he_plvio_2018}. In
normalised coordinates the line projection of
\eqn{eq:line-projection} reduces to $\boldsymbol{\ell}_n = \vn_c$
($\mathbf{K}_{\!\vL} = \mathbf{I}$), so the residual (r) is a function of the transported
moment alone, the two projectable line degrees of freedom, and never touches
position along the line:
\begin{equation}
r(\mathbf{x}, \vL_c) =
\frac{\mathbf{x}^\top \boldsymbol{\ell}_n}{\sqrt{\ell_{n,1}^2 + \ell_{n,2}^2}},
\qquad
\mathbf{r} = \begin{bmatrix} r(\mathbf{x}_s, \vL_c) \\ r(\mathbf{x}_e, \vL_c)
\end{bmatrix}.
\label{eq:residual}
\end{equation}
As noted in Section~\ref{sec:plucker}, a single observation constrains only two of the
line's four degrees of freedom; full resolution requires accumulating observations
across views with sufficient angular diversity.
The line landmark and the camera pose use different update vectors. The landmark uses
the four-dimensional perturbation $\boldsymbol{\delta}$ from \eqn{eq:orth-update},
whereas the camera pose uses the six-dimensional right perturbation
$\mathbf{T}_{cw} \leftarrow \mathbf{T}_{cw}\,\mathrm{Exp}(\boldsymbol{\xi})$,
$\boldsymbol{\xi} = [\,\boldsymbol{\varphi}^\top, \boldsymbol{\rho}^\top]^\top$, with
$\mathbf{R}_{cw} \leftarrow \mathbf{R}_{cw}\exp(\skewm{\boldsymbol{\varphi}})$ and
$\mathbf{t}_{cw} \leftarrow \mathbf{t}_{cw} + \mathbf{R}_{cw}\boldsymbol{\rho}$.
Substituting into the transport law of \eqn{eq:plk-transform} and retaining first-order
terms gives
\begin{equation}
\frac{\partial \vL_c}{\partial \boldsymbol{\xi}} =
\begin{bmatrix}
-\mathbf{R}_{cw}\skewm{\vn_w} - \skewm{\mathbf{t}_{cw}}\mathbf{R}_{cw}\skewm{\vv_w}
  & -\skewm{\vv_c}\mathbf{R}_{cw} \\
-\skewm{\vv_c}\mathbf{R}_{cw} & \mathbf{0}
\end{bmatrix}.
\label{eq:dL-dxi}
\end{equation}
The rotation block ($\boldsymbol{\varphi}$, first three columns) couples both $\vn_c$
and $\vv_c$ to the perturbation. The translation block ($\boldsymbol{\rho}$, last
three columns) acts on $\vn_c$ alone through $-\skewm{\vv_c}\mathbf{R}_{cw}$ and is
exactly zero on $\vv_c$, since translating the camera cannot change a transported
line's direction. Within the non-zero translation block, $-\skewm{\vv_c}\mathbf{R}_{cw}$
has a one-dimensional null space along $\vv_c$ itself: camera motion parallel to a
line leaves its projection unchanged. This null direction is line-specific, so a set
of parallel tracked lines shares one undetermined translation direction regardless of
how many there are. Lines spanning distinct directions instead have null spaces that
intersect only at the origin; angular diversity, not raw count, is what removes the
degeneracy.

The remaining landmark Jacobian follows by chain rule from the residual derivative,
transport law, and orthonormal update. All Jacobians are implemented analytically and
verified against central finite differences. For the single RGB-D camera evaluated here,
the camera and rig frames coincide, so no additional extrinsic mapping is required.

\subsection{Robust Weighting and Point-Line Balance}
\label{sec:weighting}

Line association is vulnerable to repetitive structure, fragmentation, and partial
occlusion, which can produce plausible but incorrect matches
\cite{gomez_plslam_2017,sun_uplslam_2025}. A single mismatched rack upright can therefore
bias the pose solve. A Cauchy robust loss~\cite{chebrolu_adaptive_2021} is applied in the
frontend pose solve; it attenuates gross outliers under iteratively reweighted least
squares without a hard cut-off.
The base information $\boldsymbol{\Omega} = (\bar{f}/\sigma_{\mathrm{px}})^2 \mathbf{I}_2$
is expressed in normalised image units and transfers across cameras of differing
resolution without retuning.

The pose solve minimises the combined objective
\begin{equation}
J(\mathbf{T}) = \underbrace{\sum_{i \in \mathcal{P}} \rho_{\text{C}}\!\left(\|\mathbf{r}^{\mathrm{p}}_i\|^2_{\boldsymbol{\Omega}_i}\right)}_{\text{point reprojection}}
\;+\;
\underbrace{w_\ell\,\lambda(N_{\mathrm{p}}) \sum_{j \in \mathcal{L}} \rho_{\text{C}}\!\left(\|\mathbf{r}^{\ell}_j\|^2_{\boldsymbol{\Omega}_j}\right)}_{\text{adaptive line term}},
\label{eq:objective}
\end{equation}
where $\mathcal{P}, \mathcal{L}$ index point and line observations, $w_\ell$ is a
fixed base weight, and $\rho_{\text{C}}$ denotes the Cauchy robust loss function. The factor $\lambda$ targets the motivating deployment scenario
directly:
\begin{equation}
\lambda(N_{\mathrm{p}}) = \mathrm{clamp}\!\left(\frac{N_{\mathrm{ref}}}{N_{\mathrm{p}}},\,
\lambda_{\min},\, \lambda_{\max}\right),
\label{eq:adaptive}
\end{equation}
with $N_{\mathrm{p}}$ the surviving point-observation count and $N_{\mathrm{ref}}$ a
reference count. In a well-textured scene, $\lambda$ falls to its floor and the solve
remains point-dominated. As point observations become sparse, $\lambda$ rises and the
line term assumes a larger share of the estimation burden, with the clamp bounding this
transfer so that a small number of lines cannot dominate the solve outright. The
implementation additionally tightens $\lambda_{\max}$ conditionally, based on a
landmark's track length and residual quality; the form in \eqn{eq:adaptive} presents the
point-scarcity term that drives this behaviour and omits these secondary caps.

\subsection{Back-End Bundle Adjustment}
\label{sec:backend}

Line landmarks are jointly refined with camera poses and point landmarks in the
sliding-window bundle adjustment. The original cuVSLAM back-end structure is retained
and extended with line-landmark states; the resulting sparse pose-landmark system is
solved by Schur complement~\cite{triggs_bundle_2000}. Because the orthonormal representation is minimal, each
line contributes a non-singular $4 \times 4$ diagonal block to $\mathbf{H}_{mm}$,
exactly analogous to a point's $3\times3$ block, marginalised by the same block-diagonal
inversion; the raw Pl\"{u}cker form would instead give $6\times6$ blocks with rank
 deficiency two, requiring explicit constraint handling. Minimality is therefore what
 makes line marginalisation numerically sound without changing the Schur-complement
 elimination structure. The
back end applies a Student's-$t$ robust loss~\cite{aravkin_student_2011} to both point and line residuals rather
differently from the frontend's Cauchy loss: the two solves address
different failure modes. The frontend rejects gross single-frame association errors
before they enter the pose estimate, while the back end's heavier-tailed loss down-weights
landmarks whose accumulated residual across many observations indicates a persistent,
rather than one-off, error.

\section{Implementation}
\label{sec:impl}

\subsection{Detection, Reconstruction and Association}
\label{sec:detection}

Detection sits behind an interchangeable detector interface, so the choice of detector
is a configuration decision rather than an architectural one. LSD~\cite{gioi_lsd_2012}
is the default used for all results reported here. Detections are filtered by length
and image-border artifacts, with the minimum length scaled to image diagonal, and are
capped at the strongest responses.

On the RGB-D path, depth is sampled along each segment and a Pl\"{u}cker line formed
directly by back-projection, with a silhouette filter rejecting segments straddling a
depth discontinuity, since such edges are viewpoint-dependent occlusion boundaries
rather than physical structure. Observations are then associated against entries in the
persistent line-landmark registry using LBD descriptors~\cite{zhang_lbd_2013} under a
geometric prior:
each active landmark is projected via \eqns{eq:plk-transform}{eq:line-projection},
and only candidates within an angular and perpendicular distance gate of that
prediction are considered, suppressing the repetitive-structure mismatches that dominate
failures in tracking environments. Survivors are matched by Hamming distance under a
ratio test. Re-associated entries are instantiated in the active optimization state,
whereas entries that leave the sliding window remain in the registry until a bounded
unobserved lifetime expires; the registry is capped in number to bound memory and solve
cost.

\subsection{Robust Multi-View Line Fusion}
\label{sec:fusion}

A landmark's 3D state is not fixed at initialization; it is refined on every
re-observation. The landmark is represented as an infinite line plus a 1D extent, an
anchor point, a unit direction, and an arc-length interval $[s_{\min}, s_{\max}]$,
rather than as two stored endpoints. Endpoints are a line's least observable
property: a detector routinely truncates or extends a segment from one frame to the
next as occlusion or contrast changes, so endpoint motion reflects measurement noise
rather than scene change.

Each observation is reconstructed independently from sampled depth and appended to a
per-landmark history buffer, with a widest-parallax-preserving eviction policy once
the buffer is full: among buffered observations, the eviction candidate is the one
whose removal least reduces the spread of viewing angles subtending the landmark,
preserving the baseline diversity that constrains the line's direction.

Fusing the buffer into a single endpoint estimate is performed with a fixed-fraction,
residual-ranked trimmed mean in the tradition of the $\alpha$-trimmed mean
filter~\cite{bednar_alphatrimmed_1984}. The buffered estimates are ranked by distance
to the pool mean, the worst-agreeing fraction is discarded, and the remainder is
averaged. The resulting closed-form estimator limits the influence of mismatched
associations and unreliable depth-based reconstructions without requiring a per-landmark
optimization loop, while preserving a continuous estimate as additional agreeing
observations are accumulated.

%
%

\section{Experiments}
\label{sec:experiments}

\subsection{Setup}
\label{sec:setup}

cuVSLAM's RGB-D odometry mode is evaluated by comparing the unmodified point-only pipeline
against the line-augmented pipeline with every other parameter fixed. The line path is
additive, so disabling it collapses the estimator to the point-only baseline and each
comparison isolates the effect of line features.

We use two datasets. The first is the TUM RGB-D freiburg3
collection~\cite{sturm_tum_2012}, all 22 sequences, whose structure and texture matrix
spans the scene properties this work targets, from textureless structured surfaces to
unstructured textured ones. The second is the Replica
dataset~\cite{straub_replica_2019}, all 8 room and office sequences of rendered RGB-D 
frames with ground-truth poses, 2000 frames per sequence.
Replica carries no sensor noise or depth artefacts, so scene geometry is the only
variable that separates its sequences.

We associate estimated trajectories to ground truth by nearest timestamp
($\Delta t \le 0.02$\,s) and align them with a rigid $\mathrm{SE}(3)$ Umeyama
transform~\cite{umeyama_1991} using evo~\cite{grupp_evo_2017}. No scale
correction is applied, since RGB-D depth is metric and any scale error therefore belongs to the
estimator that produced it. Absolute pose error (APE) RMSE is reported over the aligned
trajectory, in translation and rotation, the translation component being the absolute
trajectory error of the TUM protocol, and one-frame relative pose error (RPE) RMSE to
separate accumulated drift from frame-to-frame precision. Every sequence runs to
completion in both configurations, so all errors are computed over full trajectories.

\begin{figure}[!t]
\centering
\includegraphics[width=\columnwidth]{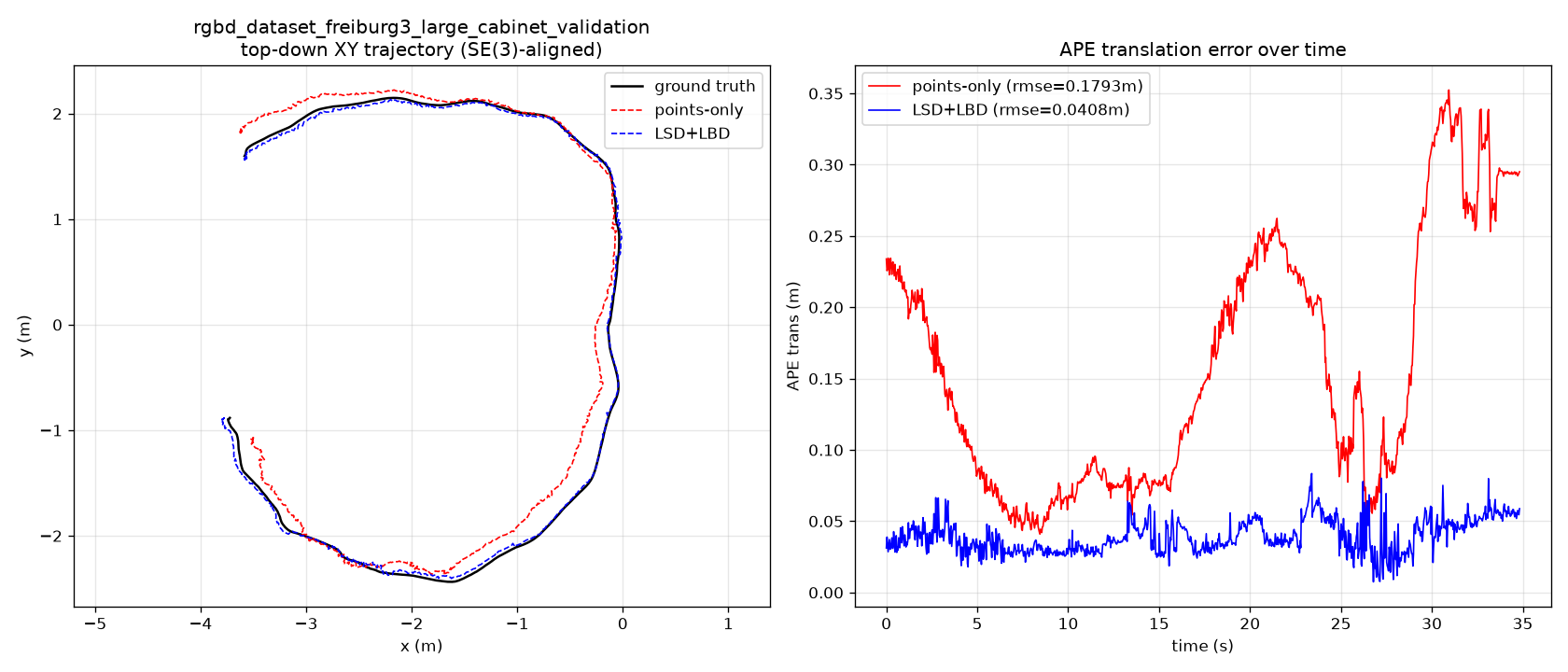}
\caption{TUM RGB-D large\_cabinet\_validation: top-down trajectory (left) and APE
translation over time (right). Points-only (red) accumulates drift to 0.18\,m RMSE, while
points+lines (blue) holds to 0.04\,m by using cabinet edge lines as persistent drift
anchors.}
\label{fig:tum-traj}
\end{figure}

\begin{figure}[b]
\centering
\begin{tabular}{@{}c@{\,}c@{}}
\includegraphics[width=0.50\columnwidth]{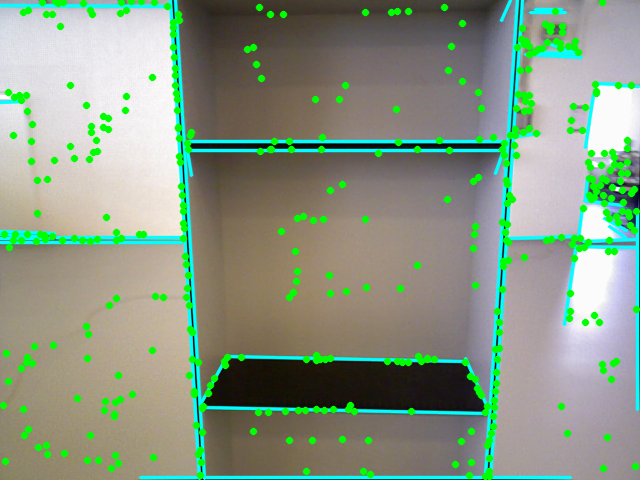} &
\includegraphics[width=0.50\columnwidth]{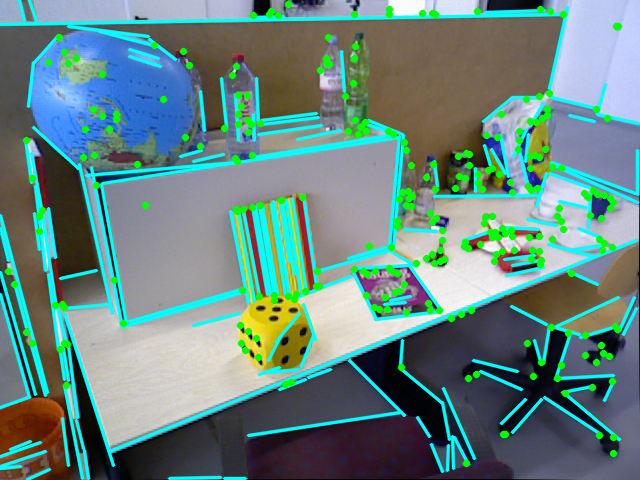}
\end{tabular}
\caption{LSD line detections (cyan) and point features (green) on TUM RGB-D frames. Left,
large\_cabinet\_validation: the flat, low-texture cabinet interior starves point
detection while shelf and frame edges still yield long, well-localised lines. Right,
long\_office\_household: desk and monitor edges give clean lines alongside dense
point coverage, the case where both modalities already agree.}
\label{fig:tum-lines}
\end{figure}

\subsection{TUM RGB-D}
\label{sec:tum-results}

Line features improve 19 of the 22 sequences (Table~\ref{tab:tum-full22}). Mean APE
translation RMSE improves from \tumAllLinesOff\,m to \tumAllLinesOn\,m (\tumAllDelta) and
the median from \tumAllMedianLinesOff\,m to \tumAllMedianLinesOn\,m (\tumAllMedianDelta).
Mean APE rotation RMSE improves from 16.72\,deg to 15.00\,deg and the median from
8.84\,deg to 5.33\,deg. The median gain exceeding the mean shows the improvement is broad
rather than carried by a few sequences.

Gains concentrate where extended, high-contrast edges persist across viewpoint change: the
cabinet sequences (large\_cabinet \largeCabinetDelta), the textured sequences
(nostructure\_texture\_far \nostructFarDelta), and the dynamic walking sequences,
whose rigid background remains edge-rich (walking\_static $-$88.6\%).
Figs.~\ref{fig:tum-traj} and~\ref{fig:tum-lines} show the bounded drift on
large\_cabinet\_validation and the detections that produce it.

\begin{table*}[t]
\centering
\small
\caption{TUM RGB-D, all 22 sequences. APE translation RMSE (m), points-only versus
points+lines. Better configuration per sequence in bold.}
\label{tab:tum-full22}
\begin{tabular}{lccc}
\toprule
\textbf{Scene} & \textbf{Points-only} & \textbf{Points+Lines} & $\mathbf{\Delta}$ \\
\midrule
cabinet & 0.2673 & \textbf{0.1325} & $-50.4\%$ \\
large\_cabinet & 0.2592 & \textbf{0.0538} & $-79.2\%$ \\
large\_cabinet\_validation & 0.1793 & \textbf{0.0408} & $-77.3\%$ \\
long\_office\_household & 0.0767 & \textbf{0.0265} & $-65.5\%$ \\
nostructure\_notexture\_far & 0.4358 & \textbf{0.2368} & $-45.7\%$ \\
nostructure\_notexture\_near\_withloop & \textbf{0.3889} & 0.7386 & $+89.9\%$ \\
nostructure\_texture\_far & 0.2652 & \textbf{0.0686} & $-74.1\%$ \\
nostructure\_texture\_near\_withloop & 0.0507 & \textbf{0.0386} & $-24.0\%$ \\
structure\_notexture\_far & 0.0642 & \textbf{0.0623} & $-3.0\%$ \\
structure\_notexture\_near & \textbf{0.0607} & 0.0970 & $+59.6\%$ \\
structure\_texture\_far & 0.0296 & \textbf{0.0109} & $-63.0\%$ \\
structure\_texture\_near & 0.0203 & \textbf{0.0112} & $-44.9\%$ \\
sitting\_halfsphere & 0.1312 & \textbf{0.0484} & $-63.1\%$ \\
sitting\_rpy & 0.0467 & \textbf{0.0342} & $-26.9\%$ \\
sitting\_static & \textbf{0.0104} & 0.0161 & $+54.6\%$ \\
sitting\_xyz & 0.0411 & \textbf{0.0189} & $-54.1\%$ \\
sitting\_xyz\_validation & 0.0282 & \textbf{0.0240} & $-14.7\%$ \\
teddy & 0.2898 & \textbf{0.0689} & $-76.2\%$ \\
walking\_halfsphere & 0.4575 & \textbf{0.3320} & $-27.4\%$ \\
walking\_rpy & 0.7629 & \textbf{0.3698} & $-51.5\%$ \\
walking\_static & 0.1560 & \textbf{0.0178} & $-88.6\%$ \\
walking\_xyz & 0.2895 & \textbf{0.2033} & $-29.8\%$ \\
\midrule
\textbf{Mean} & \textbf{0.1960} & \textbf{0.1205} & \textbf{$-38.5\%$} \\
\bottomrule
\end{tabular}

\end{table*}

Ten of the 22 sequences form the focus subset used for the published cuVSLAM Odom
comparison, which excludes the low-texture and dynamic sequences. On this subset mean APE
translation RMSE improves from \tumSubsetLinesOff\,m to \tumSubsetLinesOn\,m
(\tumSubsetDelta) and every sequence improves.
Table~\ref{tab:tum-10scene-baseline} places the subset against published results: the
points-only mean is consistent with the 0.11\,m reported for cuVSLAM Odom, confirming the
evaluation protocol, and the line-augmented pipeline reaches \tumSubsetLinesOn\,m, below the
published ORB-SLAM3 value of 0.067\,m.

\begin{table}[b]
\centering
\small
\caption{TUM RGB-D, ten-sequence subset means. APE translation RMSE (m) for published
methods and for the two configurations.}
\label{tab:tum-10scene-baseline}
\begin{tabular}{lc}
\toprule
\textbf{Method} & \textbf{RMSE APE (m)} \\
\midrule
ORB-SLAM3 & 0.067 \\
DPVO & 0.80 \\
cuVSLAM Odom (paper baseline) & 0.11 \\
\midrule
cuVSLAM Points-only (our eval) & 0.1112 \\
cuVSLAM Points + LSD+LBD & \textbf{0.0357} \\
Improvement & \textbf{$-67.9\%$} \\
\bottomrule
\end{tabular}

\end{table}

\begin{figure}[b]
\centering
\begin{tabular}{@{}c@{\,}c@{}}
\includegraphics[width=0.50\columnwidth]{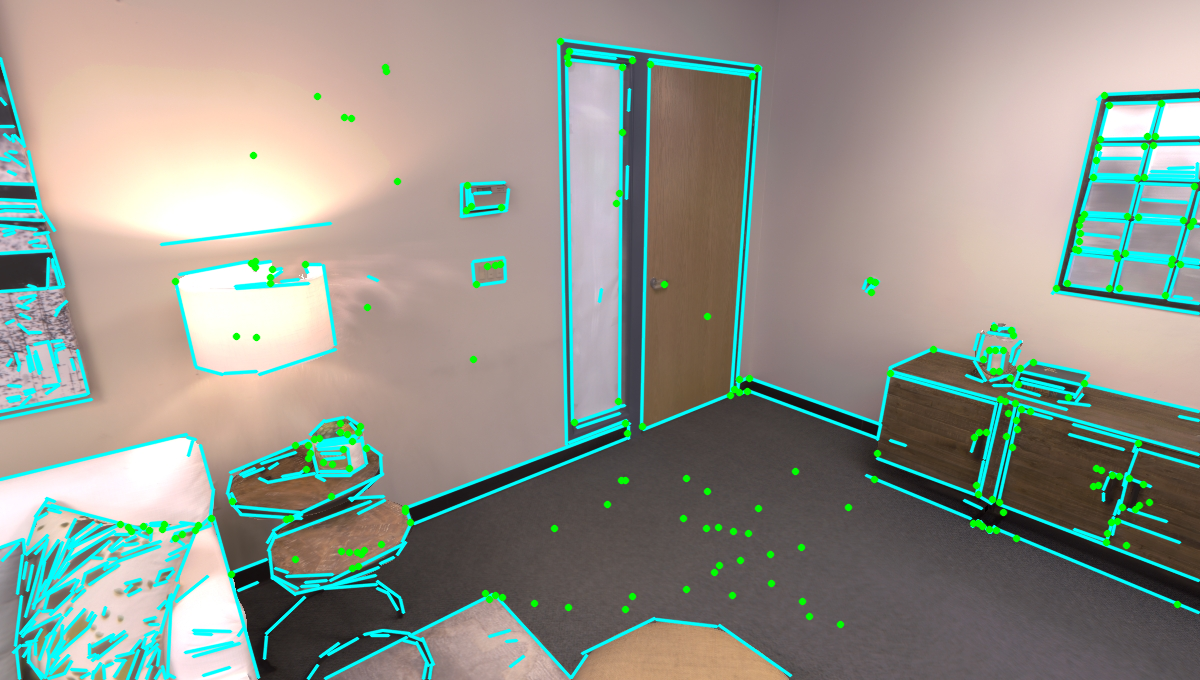} &
\includegraphics[width=0.50\columnwidth]{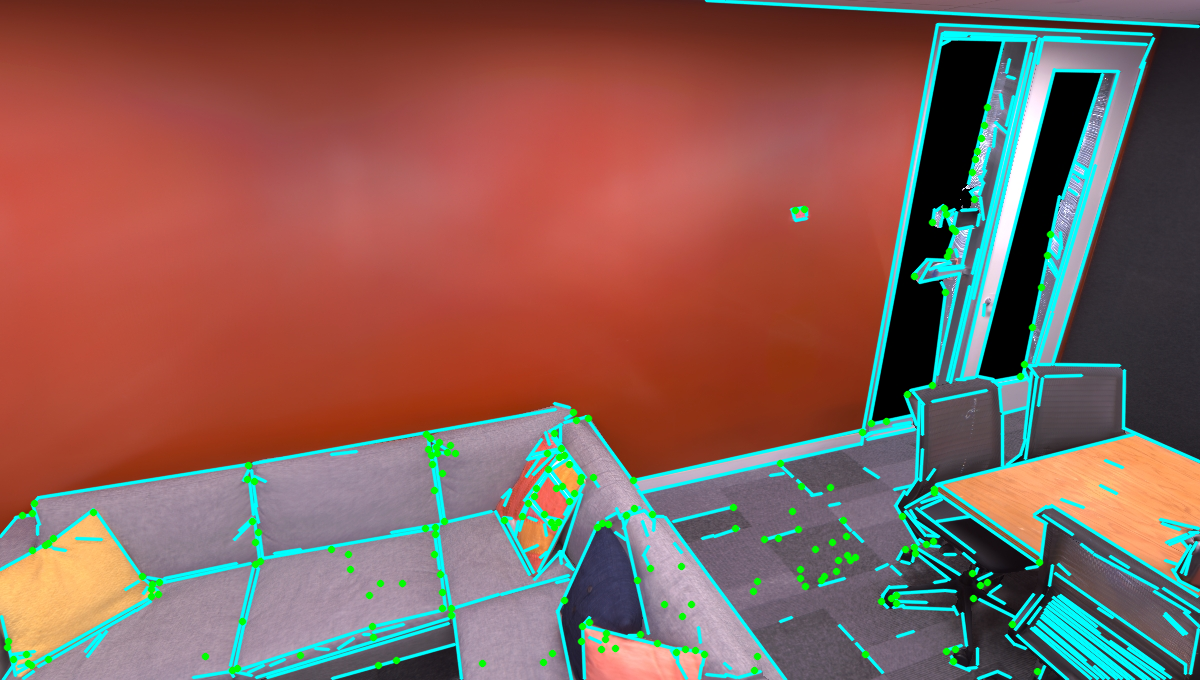}
\end{tabular}
\caption{LSD line detections (cyan) and point features (green) on Replica frames. Left,
room0: door and window-shelf frames and a wall-mounted grid provide stable line
evidence over otherwise blank wall and floor. Right, office3: wall edges and
furniture outlines anchor lateral drift where the wall carries almost no point texture.}
\label{fig:replica-lines}
\end{figure}

Three sequences regress. On structure\_notexture\_near the error grows from 0.0607\,m to
0.0970\,m while the far variant of the same scene is unchanged (0.0642\,m against
0.0623\,m), and on nostructure\_notexture\_near\_withloop, already among the least accurate
of the 22, it grows from 0.3889\,m to 0.7386\,m. Both are near-range views of untextured
surfaces, the regime in which $\lambda(N_{\mathrm{p}})$ is largest. On sitting\_static the
error grows from 0.0104\,m to 0.0161\,m, a 5.7\,mm difference on a sequence whose
ground-truth trajectory is 0.259\,m long.

One-frame RPE rises on the full set, mean translation from 0.0108\,m to 0.0136\,m and mean
rotation from 0.401\,deg to 0.486\,deg, so line features trade a small amount of per-step
precision for global consistency, the appropriate direction for a front end prioritising
robustness over frame-to-frame accuracy. That cost sits entirely in the twelve sequences outside the focus subset. On the
focus subset itself RPE is level (0.0102\,m against 0.0104\,m in translation, 0.376\,deg
against 0.393\,deg in rotation).

\subsection{Replica Dataset}
\label{sec:replica-results}

\begin{table*}
\centering
\small
\caption{Replica dataset, all 8 sequences. APE translation and rotation RMSE, points-only
versus points+lines. Better configuration per sequence in bold.}
\label{tab:replica}
\begin{tabular}{lcccccc}
\toprule
\textbf{Scene} & \multicolumn{2}{c}{\textbf{Trans (m)}} & \multicolumn{2}{c}{\textbf{Rot (deg)}} & \multicolumn{2}{c}{\textbf{Improvement}} \\
& \textbf{Pts} & \textbf{Pts+Lines} & \textbf{Pts} & \textbf{Pts+Lines} & \textbf{Trans} & \textbf{Rot} \\
\midrule
room0 & 0.0129 & \textbf{0.0079} & 0.239 & \textbf{0.163} & $-38.8\%$ & $-31.8\%$ \\
room1 & 0.0124 & \textbf{0.0045} & 0.822 & \textbf{0.176} & $-63.7\%$ & $-78.6\%$ \\
room2 & \textbf{0.0093} & 0.0101 & 0.453 & \textbf{0.196} & $+8.6\%$ & $-56.7\%$ \\
office0 & 0.0173 & \textbf{0.0169} & \textbf{0.652} & 1.242 & $-2.3\%$ & $+90.5\%$ \\
office1 & 0.0103 & \textbf{0.0091} & 0.997 & \textbf{0.547} & $-11.7\%$ & $-45.1\%$ \\
office2 & 0.0159 & \textbf{0.0054} & 0.509 & \textbf{0.194} & $-66.0\%$ & $-61.9\%$ \\
office3 & 0.0140 & \textbf{0.0040} & 0.235 & \textbf{0.138} & $-71.4\%$ & $-41.3\%$ \\
office4 & 0.0342 & \textbf{0.0303} & 1.022 & \textbf{0.956} & $-11.4\%$ & $-6.4\%$ \\
\midrule
\textbf{Mean} & \textbf{0.0158} & \textbf{0.0110} & \textbf{0.616} & \textbf{0.451} & \textbf{$-30.1\%$} & \textbf{$-26.8\%$} \\
\textbf{Median} & \textbf{0.0135} & \textbf{0.0085} & \textbf{0.580} & \textbf{0.195} & \textbf{$-38.9\%$} & \textbf{$-66.4\%$} \\
\bottomrule
\end{tabular}

\end{table*}

Across all eight Replica sequences, mean APE translation RMSE improves from 0.0158\,m to
0.0110\,m (\replicaMeanTransDelta) and the median from 0.0135\,m to 0.0085\,m
(\replicaMedianTransDelta); mean APE rotation RMSE improves from 0.616\,deg to 0.451\,deg
(\replicaMeanRotDelta) and the median from 0.580\,deg to 0.195\,deg
(\replicaMedianRotDelta). Seven of eight sequences improve in each metric
(Table~\ref{tab:replica}). The median rotation gain is the largest of the
four aggregates because the two sequences with the worst points-only rotation drift, room1
at 0.822\,deg and office1 at 0.997\,deg, are corrected almost entirely.

The largest gains are on the structure-rich office sequences, office3 by 71\% in
translation and office2 by 66\%, where extended wall, door, and furniture edges survive on
surfaces carrying almost no point texture (Fig.~\ref{fig:replica-lines}). Two sequences
show isolated regressions, indicating that line detection can introduce noise in certain
contexts; on room2, RPE translation RMSE rises from 0.0007\,m to 0.0042\,m while its 
median falls, and on office0, APE rotation regresses from 0.652\,deg to 1.242\,deg at 
unchanged translation accuracy.

Taken across the 30 sequences of both datasets, translation accuracy improves on 26 and no
sequence loses tracking, so line features gain accuracy and robustness together, with the
residual cost confined to per-step precision.

\section{Discussion}
\label{sec:discussion}

Line features deliver their gain in the error mode that governs autonomous mobile robot
(AMR) operation: accumulated drift, which sets how far the platform travels before it
requires relocalisation. Translation accuracy improves on 26 of the 30 sequences and rotation
accuracy on 7 of 8 Replica sequences, with no tracking failure in any configuration.
Accuracy and robustness therefore improve together; an estimator that trades occasional
track loss for accuracy is not deployable at fleet scale.

The geometry of the constraint explains where the gain comes from. The translation block of
\eqn{eq:dL-dxi} leaves motion along a line unobserved, and parallel lines share that null
direction however many are tracked (Section~\ref{sec:residual}), so lines contribute in
proportion to how well their directions complement the angular distribution of the surviving
point observations. The Replica office sequences supply that complement through wall, door
and furniture edges on surfaces carrying almost no point texture, and they show the largest
gains of the 30 sequences. Structured indoor environments supply this evidence densely,
which establishes the operational case for the modality.

The adaptive weight of \eqn{eq:adaptive} is self-limiting, sitting at its floor throughout
well-textured scenes such as most of TUM RGB-D, so the feature need not be restricted to
known feature-poor routes. All runs completed without tracking failure on the two evaluated
datasets. Within these experiments, the observed trade-off was therefore in accuracy rather
than track completion, and downstream recovery of a bounded accuracy change is preferable
to recovery after a lost track.

Persistent line landmarks bound the accumulated error, and they do so through a measurement
path that is deliberately conservative. The perpendicular endpoint-to-line
residual of \eqn{eq:residual} is invariant to position along the line, so the endpoint
truncation that dominates segment instability never reaches the solver. The results indicate
that per-frame LSD re-detection introduces additional variation into the line residuals in
some sequences, which can limit short-term accuracy while improving robustness to accumulated
drift. On the full TUM set, mean one-frame RPE translation increases from 0.0108\,m to
0.0136\,m, while APE translation improves on 19 of 22 sequences (Section~\ref{sec:tum-results}).
Across both datasets, translation accuracy improves on 26 of the 30 sequences. Coupling
per-frame detection with direct line tracking,
so that a landmark's image support is propagated instead of refitted, removes this noise at
its source and is the next step for the front end.

The exceptions are three TUM
sequences and two Replica metrics, all small in absolute terms: sitting\_static shifts by
5.7\,mm on a 0.259\,m trajectory, room2 loses 0.8\,mm of translation accuracy, and
office0's rotation error rises at unchanged translation accuracy. These cases are consistent with
a failure mode in which the point-count-based weight increases line authority while repetitive-edge
associations are less reliable. Future work will explore context-aware per-observation
uncertainty, which weighs each line measurement on its own evidence rather than transferring
authority on point count alone.

The observed accuracy trade-off is therefore associated with the measurement path and
the apportionment of estimation authority between the point and line modalities, rather
than with the Pl\"{u}cker--orthonormal formulation or its Schur-complement marginalisation.

\section{Conclusion}
\label{sec:conclusion}

An extension of cuVSLAM with persistent line landmarks, trimmed-mean multi-view fusion,
and a clamped adaptive point-line weight has been presented for structured indoor
environments. Lines are incorporated as additional residual blocks, leaving the point
pipeline unchanged when the line terms are disabled. The Pl\"{u}cker--orthonormal
parameterisation provides minimal $4 \times 4$ landmark blocks that are marginalised by
the existing Schur complement without the rank deficiency of raw Pl\"{u}cker coordinates.

Under $\mathrm{SE}(3)$ Umeyama alignment with no scale correction, mean APE translation
RMSE is reduced by \tumAllDeltaMag\ over the 22 TUM RGB-D sequences and by
\replicaMeanTransDeltaMag\ over the 8 Replica sequences. Translation improves on 19 of
22 TUM sequences and 7 of 8 Replica sequences, while no tracking failures are observed.
On the ten-sequence focus subset, the line-augmented pipeline reaches
\tumSubsetLinesOn\,m, below the published cuVSLAM Odom and ORB-SLAM3 values. The gain is
concentrated in accumulated drift, with a small rise in one-frame RPE.

The improvement is not unconditional. Regressions occur in four of thirty sequences, in low-texture, untextured near-range
scenarios where repetitive-edge associations become unreliable; the point-count-based
weighting rule increases line authority precisely in these regimes, amplifying rather than
suppressing detection noise. Future work should explore maximising the line signal through
more accurate line detection and matching to improve robustness in such challenging cases.


\section*{Acknowledgment}
This work is an extension of cuVSLAM~\cite{korovko_cuvslam_2025}. We thank the cuVSLAM
team, whose CUDA-accelerated point-feature front end, sliding-window bundle adjustment
and solver architecture this line path builds on directly. We also thank the authors of
PL-VINS~\cite{fu_plvins_2020}, whose open-source line geometry and line-factor
implementation informed the formulation.

{\footnotesize
\bibliographystyle{IEEEtran}
\bibliography{LineFeaturesCuVSLAM}
}

\end{document}